\documentclass{article}

\usepackage{arxiv}

\usepackage[utf8]{inputenc} 
\usepackage[T1]{fontenc}    
\usepackage{textcomp}       

\usepackage{hyperref}       
\usepackage{url}            
\usepackage{booktabs}       
\usepackage{amsfonts}       

\usepackage{graphicx}
\usepackage{tikz}
\usepackage{pgfplots}
\pgfplotsset{compat=1.17}

\usepackage{amsmath}

\usepackage{subfig}
\usepackage{color}
\usepackage{float}
\usepackage{tabularx}
\usepackage{appendix}
\newcolumntype{C}{>{\centering\arraybackslash}X}

\title{Language Detection by Means of the Minkowski Norm: Identification Through Character Bigrams and Frequency Analysis}

\author{
 Paul-Andrei Pog\u{a}cean \\
  Babeș-Bolyai University \\
  \texttt{paul.pogacean@stud.ubbcluj.ro} \\
   \And
 Sanda-Maria Avram \\
  Babeș-Bolyai University \\
  \texttt{sanda.avram@ubbcluj.ro} \\
}

\begin{document}
\maketitle


 \begin{abstract}
The debate surrounding language identification has gained renewed attention in recent years, especially with the rapid evolution of AI-powered language models. However, the non-AI-based approaches to language identification have been overshadowed. This research explores a mathematical implementation of an algorithm for language determinism by leveraging monograms and bigrams frequency rankings derived from established linguistic research. The datasets used comprise texts varying in length, historical period, and genre, including short stories, fairy tales, and poems. Despite these variations, the method achieves over 80\% accuracy on texts shorter than 150 characters and reaches 100\% accuracy for longer texts. These results demonstrate that classical frequency-based approaches remain effective and scalable alternatives to AI-driven models for language detection.
\end{abstract} 

\keywords{n-grams, monograms, bigrams, language detection, statistical language identification}

\bigskip

\section{Introduction} \label{intro}

Language identification is a fundamental task in natural language processing (NLP)~\cite{lui2012langid}, with broad applications across machine learning~\cite{joulin2017}, information retrieval~\cite{huang2003}, and text classification~\cite{cavnar1994}. Approaches to this task generally fall into two categories: AI-based methods, such as deep learning and neural networks~\cite{brown2020}, and non-AI-based, statistical methods~\cite{mcnamee2005}. While AI-based techniques dominate current research, they often require large training datasets, substantial computational resources, and complex model management.

In contrast, statistical methods leveraging character frequencies, diacritics, and bigrams provide lightweight and accessible alternatives. Prior studies have investigated n-gram classification and character frequency analysis~\cite{norvig2012}. However, there is limited research on diacritic pattern utilization, and to our knowledge, no work has combined these features systematically. This paper proposes a novel approach that integrates these non-AI techniques to improve language identification accuracy, particularly for texts of at least 150 characters, while maintaining robustness in challenging scenarios.

We evaluate our method using four datasets: the ROST (Romanian Stories and Tales) dataset~\cite{sanda2022kaggle}, a multilingual collection of fairy tales~\cite{povesti2024}, the OPUS Multilingual Parallel Corpus~\cite{opus}, and a corpus of poems~\cite{dataset4}. These datasets encompass a variety of text lengths, genres, and historical periods, providing a comprehensive and diverse testbed for language identification.

The remainder of this paper is organized as follows:
\begin{itemize}
    \item Section~\ref{related} reviews related work in language identification.
    \item Section~\ref{methods} describes our methodology.
    \item Section~\ref{results} presents experimental results.
    \item Section~\ref{conclusion} concludes and outlines future research directions.
\end{itemize}

\section{Related Work} \label{related}
\subsection{Overview}
Language detection is a longstanding research area in computational linguistics~\cite{lui2012langid}, with early work focused on statistical and rule-based approaches~\cite{cavnar1994,mcnamee2005}. The advent of artificial intelligence (AI) has since transformed the field, enabling more robust and scalable solutions~\cite{joulin2017}. Today, language identification underpins applications ranging from machine translation to information retrieval.

\subsection{AI-Based Approaches}
Modern language identification systems often rely on deep learning and large language models. Google Translate and DeepL use neural architectures trained on massive multilingual corpora, achieving high accuracy even on ambiguous or mixed-language input~\cite{joulin2017,deepl}. OpenAI's GPT models demonstrate strong language identification as a byproduct of their text understanding capabilities~\cite{brown2020}. Facebook's FastText~\cite{joulin2017} offers a lightweight, efficient solution, particularly suited for real-time applications such as social media.

While these AI-driven methods are highly effective, they require substantial computational resources and large annotated datasets, which may not be feasible in all settings.

\subsection{Non-AI-Based Approaches}
Early search engines used heuristic and statistical methods, including n-gram models and URL analysis, to infer language~\cite{huang2003}. Before the rise of AI, statistical models such as n-gram analysis were the primary tools for language identification~\cite{cavnar1994}. These methods compare the frequency distributions of character sequences (monograms, bigrams, etc.) in a text to language-specific profiles~\cite{norvig2012}. Frequency-based approaches remain attractive for their simplicity, interpretability, and low computational overhead. However, their accuracy can be challenged by short texts, code-switching (i.e., alternating between two or more languages or language varieties within a single conversation or even within a sentence), or closely related languages~\cite{evans2024,lewand2000,mitrea2012,trost2024,trost2005,trost2024turkish,zuidema2009}.

Rule-based systems, relying on handcrafted linguistic rules and dictionaries, have also been employed, especially for languages with distinctive orthographic features~\cite{mcnamee2005}. 

\subsection{Mathematical Tools: The Minkowski Norm}
Recent research has explored various distance metrics to compare language profiles. The Minkowski (or \textbf{p-norm}) distance, particularly the Manhattan norm, obtained when \( p = 1 \), is widely used for measuring differences between frequency vectors. This metric is especially suited for high-dimensional, sparse data such as character n-gram distributions.

\subsection{Datasets} \label{datasets}
As we already stated in Section~\ref{intro}, we evaluate our approach using four datasets, which are described in detail in Table~\ref{tab:datasets}.
\begin{table}[H]
    \centering
    \renewcommand{\arraystretch}{1.1}
    \begin{tabular}{|p{0.2\textwidth}|p{0.75\textwidth}|}
        \hline
        \textbf{Dataset} & \textbf{Description} \\
        \hline
        \textbf{FTDS} (Fairy Tales Data Set) &
        \textit{Content:} Romanian fairy tales and children's stories. \newline
        \textit{Language:} Romanian. \newline
        \textit{Size:} 100+ stories, varying in length and style. \newline
        \textit{Use:} Study of narrative structure, vocabulary, and syntax in literary Romanian. \newline
        \textit{Source:} \url{https://www.povesti-pentru-copii.com} \\
        \hline
        \textbf{LDDS} (Language Determinism Data Set) &
        \textit{Content:} Multilingual texts across several languages (e.g., English, Romanian, German, French). \newline
        \textit{Language:} Multilingual. \newline
        \textit{Size:} ~1,000+ samples per language. \newline
        \textit{Use:} Comparative linguistic analysis and testing of language determinism. \newline
        \textit{Source:} \url{https://github.com/PogaceanPaul/DataSetLanguage/tree/main} \\
        \hline
        \textbf{OPUS} (Multilingual Parallel Corpus) &
        \textit{Content:} Sentence-aligned parallel corpora for multiple language pairs (e.g., English–Romanian). \newline
        \textit{Language:} Multilingual. \newline
        \textit{Size:} Millions of aligned sentence pairs (e.g., TED Talks, GNOME, Europarl). \newline
        \textit{Use:} Cross-linguistic comparison, machine translation, and language modeling. \newline
        \textit{Source:} \url{http://opus.nlpl.eu/} \\
        \hline
        \textbf{ROST} (Romanian Stories and Texts) &
        \textit{Content:} Romanian literary texts (classical and contemporary). \newline
        \textit{Language:} Romanian. \newline
        \textit{Size:} Several hundred text segments grouped by author or theme. \newline
        \textit{Use:} Stylistic, lexical, and syntactic analysis of Romanian. \newline
        \textit{Source:} \url{https://doi.org/10.34740/KAGGLE/DS/2545712} \\
        \hline
    \end{tabular}
    \caption{Overview of the datasets used in this project}
    \label{tab:datasets}
\end{table}


\subsection{Summary and Gap}

While AI-based methods dominate current research, lightweight statistical approaches remain valuable for resource-constrained environments and interpretable applications. However, limited work has systematically combined character frequency, bigram analysis, and diacritic patterns for language identification. Our work addresses this gap by integrating these features and evaluating their effectiveness across diverse datasets.

\section{Methodology and Algorithm Development}\label{methods}

This section details the iterative development of our non-AI language identification algorithm, emphasizing the mathematical and empirical rationale for each design choice. We will use C++ as the programming language for our implementation because it offers faster execution and better performance than other high-level languages like Python~\cite{202012.0516}, especially when processing large datasets. This choice supports an efficient implementation capable of handling high data volumes.

\subsection{Terminology and Framework}

We introduce key terms to clarify our approach:
\begin{description}
    \item[\textbf{Corpora}] Collections of texts in a given language, spanning diverse styles, authors, and historical periods.
    \item[\textbf{Frequency List}] A ranked list of the $n$ most frequent characters in a language, derived from a representative corpus.
    \item[\textbf{Frequency Distribution}] The observed frequencies of characters within a specific text.
    \item[\textbf{Language Detection}] The process of applying multiple scoring strategies to attribute a text to the most likely language, as detailed below.
    \item[\textbf{Formal Texts}] Texts with standard orthography, including diacritics and correct spelling.
    \item[\textbf{Informal Texts}] Texts with non-standard spelling, contractions, or missing diacritics.
    \item[\textbf{Noisy Texts}] Texts containing typographical, grammatical, or semantic errors.
    \item[\textbf{Noise Model}] A noise transformation $\tau: T \to T'$ that can involve one of the following operations:
\begin{enumerate}
    \item \textbf{Vowel deletion}: Removal of a vowel $v \in V$ from the text.
    \item \textbf{Vowel substitution}: Replacement of a vowel $v \in V$ with a consonant $c \in C$.
    \item \textbf{Symbol substitution}: Replacement of an alphanumeric character $a \in A$ with a non-alphanumeric symbol.
\end{enumerate}
\end{description}

\subsection{Mathematical Foundation}

Character frequency distributions can be represented as vectors in an $n$-dimensional space, where $n$ is the number of unique characters considered. The distance between a text's character frequency vector and each language's reference vector is computed using a custom norm inspired by the Manhattan ($\ell_1$) norm. This enables robust comparison across languages.

While monogram (single character) frequency analysis is often sufficient, it may be inconclusive for languages with similar character distributions. In such cases, bigram (character pair) analysis provides additional discriminatory power~\cite{mitrea2012}.

\subsection{Algorithm Development: Iterative Refinement}

Our algorithm evolved through four major versions, each addressing limitations identified in the previous iteration.

\subsubsection{Version 1: Top-5 Character Frequency Patterns}

The initial version hypothesized that a language could be identified by analyzing the top five most frequent characters in a text. We normalized character frequencies and compared patterns across Romanian, German, and English texts from the FTDS dataset. Table~\ref{tab:algorithm_results} summarizes these results.

\begin{table}[ht]
    \centering
    \begin{tabular}{|l|l|c|c|c|c|c|r|r|r|r|r|}
        \hline
        \textbf{Title} & \textbf{Language} & \multicolumn{5}{|c|}{\textbf{Top 5 letters}} & \multicolumn{5}{|c|}{\textbf{Percentages}} \\
        \hline
        Snow White & German & e & n & i & s & a & 13.50 & 8.60 & 6.45 & 4.65 & 4.60 \\
        \hline
        Snow White & Romanian & e & i & a & \u{a} & r & 7.67 & 7.64 & 7.55 & 5.53 & 4.72 \\
        \hline
        Snow White & English & e & t & a & o & h & 9.84 & 6.66 & 6.35 & 5.92 & 5.84 \\
        \hline
        Hansel and Gretel & German & e & n & i & s & r & 12.97 & 8.36 & 5.97 & 4.89 & 4.86 \\
        \hline
        Hansel and Gretel & Romanian & e & i & a & r & \u{a} & 8.49 & 7.77 & 6.87 & 5.13 & 4.82 \\
        \hline
        Hansel and Gretel & English & e & t & a & o & h & 10.56 & 6.82 & 6.50 & 5.71 & 5.34 \\
        \hline
        Little Red Riding Hood & German & e & n & r & i & t & 10.86 & 6.97 & 5.07 & 5.00 & 4.94 \\
        \hline
        Little Red Riding Hood & Romanian & i & a & e & u & c & 8.70 & 7.48 & 7.21 & 5.23 & 4.44 \\
        \hline
        Little Red Riding Hood & English & e & t & o & h & a & 9.89 & 7.1 & 6.49 & 5.72 & 5.41 \\
        \hline
    \end{tabular}
    \caption{Results of the first version of the algorithm}
    \label{tab:algorithm_results} 
\end{table}

\textbf{Key findings:}
\begin{itemize}
    \item English texts: The most frequent character stands out by about 0.5\% over the next most frequent.
    \item German texts: A sharp drop after the fourth most frequent character.
    \item Romanian texts: A smoother, more gradual frequency decline.
    \item The top five characters alone are insufficient for consistent identification across all texts and languages.
\end{itemize}

\textbf{Theoretical Justification:}
We can notice that a frequency list of length 25 is statistically sufficient to uniquely identify languages, given the large number of possible permutations ($\geq 25!$) relative to the number of world languages (7151 languages worldwide, estimated as of January 19, 2024).

\subsubsection{Version 2: Scoring with Top-10 Character Frequencies}

Building on these insights, the second version:
\begin{enumerate}
    \item Constructed a reference mapping of the 25 most frequent characters per language, assigning scores from 25 (most frequent) to 1.
    \item Extracted the top 10 characters from the input text.
    \item Calculated a language score by comparing these characters to the reference mapping.
\end{enumerate}

\begin{figure}[H]
\includegraphics[width=1\textwidth]{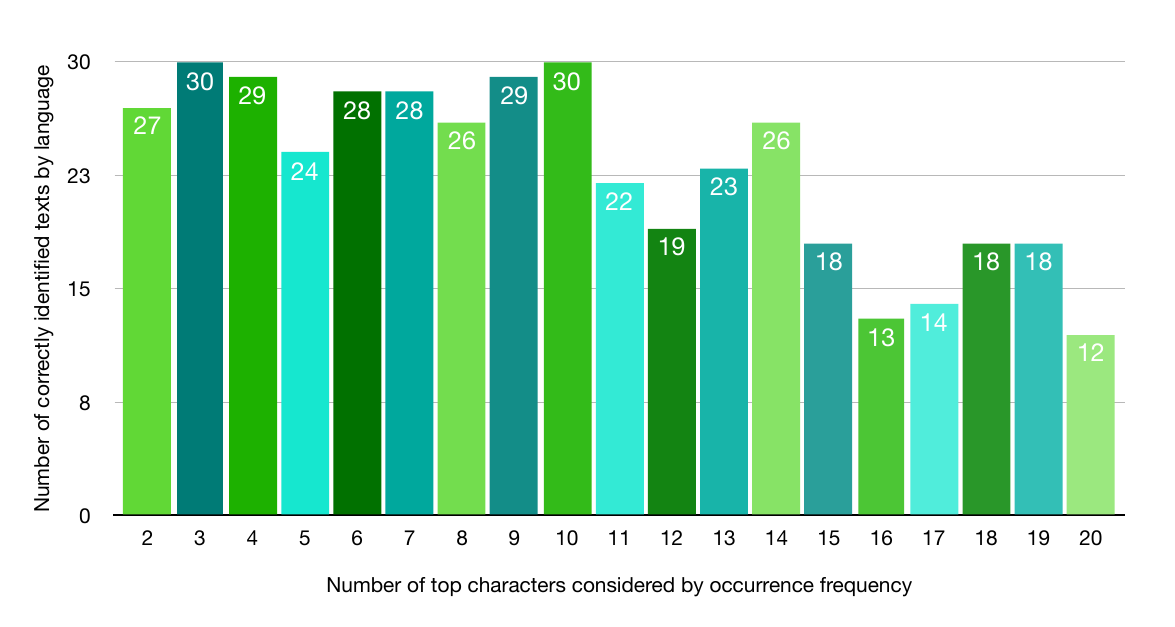}
\caption{Algorithm accuracy for $m \in [2, 20]$}
\label{fig:proof-of10}
\end{figure}

Empirical testing across a range of texts (with and without diacritics) established that a threshold of 10 characters balances accuracy and robustness (see Figure~\ref{fig:proof-of10}).

The scoring function is based on the Minkowski norm, with the highest score determining the detected language.

\subsubsection{Version 3: Expanding Languages and Robustness Analysis}

This iteration expanded the language set to include Hungarian and Turkish, and tested the algorithm on a broader range of texts and authors. Accuracy exceeded 86\% for texts over 150 characters. However, several factors were found to hinder performance:
\begin{itemize}
    \item Use of slang, typos, or non-standard forms alters frequency patterns.
    \item Short texts reduce statistical reliability.
    \item Historical or archaic texts deviate from modern language profiles.
    \item Missing diacritics, especially in Romanian, lead to misclassification.
\end{itemize}

The algorithm is most effective for formal texts with preserved diacritics. Testing confirmed that diacritic characters are key discriminators and typically appear in the top 10 for such texts.

\textbf{Mathematical Analysis of Noise Effects:}

We formalize the impact of noise (typos, missing vowels, symbol substitutions) on frequency profiles. Using a frequency ranking function and $\ell_1$ norm, we prove that noise operations systematically reduce the expected frequency score of the top characters, thus degrading identification accuracy. This can be achieved by means of probabilities. 

\textit{Proof:} Let us analyze the effect of each noise operation on the $\ell_1$ norm of the frequency vector $\mathbf{g}$, which contains frequency scores of the top 10 characters in the original text $T$.

\medskip

\noindent Define the top character set as:
\[
C_{\text{top}} = \{c \in \Sigma : f(c) \text{ is among the 10 highest in } T\}
\]
Let $\mathbf{g}$ denote the frequency score vector over $C_{\text{top}}$, and let $\mathbf{g}'$ be the corresponding vector after applying a noise operation $\tau$ to produce the noisy text $T' = \tau(T)$.

\bigskip

\noindent \textbf{1. Vowel Deletion:} \\
Consider a vowel $v \in C_{\text{top}}$ with frequency score $f(v)$. If $v$ is deleted from $T$, its contribution is removed from $\|\mathbf{g}\|_1$. If a new character $a \notin C_{\text{top}}$ enters the top 10, its score will satisfy $f(a) \leq \min_{c \in C_{\text{top}}} f(c)$. Thus, the change in the norm is at most:

\[
\Delta_1 = -f(v) + \max_{a \notin C_{\text{top}}} f(a)
\]

Since vowels typically have high frequency, and any replacement is likely to have a lower score, the expected change is negative:
\[
\mathbb{E}[\Delta_1] = \mathbb{E}[\|\mathbf{g}'\|_1 - \|\mathbf{g}\|_1 \mid \text{Vowel Deletion}] < 0
\]

\bigskip

\noindent \textbf{2. Vowel Substitution:} \\
Suppose a vowel $v$ is replaced by a consonant $c$ with score $f(c)$. Then:

\[
\Delta_2 = f(c) - f(v)
\]

Because vowels on average occur more frequently than consonants, the expected change is also negative:
\[
\mathbb{E}[\Delta_2] = \mathbb{E}[\|\mathbf{g}'\|_1 - \|\mathbf{g}\|_1 \mid \text{Vowel Substitution}] = \mathbb{E}[f(c) - f(v)] < 0
\]

\bigskip

\noindent \textbf{3. Symbol Substitution:} \\
Let $a \in C_{\text{top}}$ be replaced by a non-alphanumeric symbol. This symbol does not contribute to the frequency vector, so the total score drops by $f(a)$. If a new character $b \notin C_{\text{top}}$ enters the top 10, it has:

\[
\Delta_3 = -f(a) + \max_{b \notin C_{\text{top}}} f(b)
\]

Again, the expected result is a net decrease:
\[
\mathbb{E}[\Delta_3] = \mathbb{E}[\|\mathbf{g}'\|_1 - \|\mathbf{g}\|_1 \mid \text{Symbol Substitution}] < 0
\]

\bigskip

\noindent \textbf{Conclusion:} \\
Each noise operation $\tau$ results in a negative expected change in the $\ell_1$ norm of the frequency vector over the top 10 characters. Therefore, for a noisy text $T' = \tau(T)$:

\[
\mathbb{E}[\|\mathbf{g}'\|_1] = \|\mathbf{g}\|_1 + \mathbb{E}[\Delta] < \|\mathbf{g}\|_1
\]

Since the language detection algorithm relies on these frequency scores, this degradation can mislead the scoring function and lower detection accuracy.

\hfill $\square$

\subsubsection{Version 4: Final Algorithm}

Building on the iterative development described previously, the fourth version of our algorithm integrates two key enhancements: explicit diacritic evaluation and bigram-based scoring. The next section (\ref{results}) details these refinements, their mathematical formalization, and the empirical impact of these refinements.

\section{Final Algorithm and Results}
\label{results}

\subsection{Incorporating Diacritic Evaluation}

To address the challenge of distinguishing languages with similar character distributions, we introduce a diacritic-based bonus. If the proportion of diacritical characters in the input text exceeds a threshold percentage ($p = 5\%$ or $10\%$), a bonus of $k = 100$ or $200$ points is awarded to the corresponding language. The bonus given is meant to help us determine the language of shorter texts (fewer than 150 characters), where statistical reliability of character frequency is otherwise limited. For longer texts, the bonus is negligible relative to the overall score, preserving the integrity of the frequency-based method. In cases where the input is very short or contains quotations and proper nouns from other languages, additional heuristics need to be applied.

\subsection{Bigram-Based Scoring}

Recognizing that monogram analysis alone may not sufficiently distinguish between languages with overlapping character frequencies (e.g., Dutch vs. English, German vs. Romanian without diacritics), we incorporate a bigram-based scoring mechanism. For each language, we compile a list of the ten most frequent bigrams, as established in prior linguistic studies~\cite{mitrea2012}. Each bigram is assigned a score from 1 to 10, analogous to the ranking of the top 25 letters. The total bigram bonus is computed as the sum of the scores for each recognized bigram occurrence in the input text. By limiting the list to ten bigrams per language, we ensure bounded, interpretable contributions and prevent score inflation.

\subsection{Unified Scoring Formula}

The final language score for a text is thus computed as:
\begin{equation} 
    \|(X, F)\| = \sum_{i=1}^m |X_i| + \sum_{i=1}^m |F_i \cdot i| + \text{bonus}
    \label{finaleq}
\end{equation}
where $X$ and $F$ are the character and bigram frequency vectors, respectively, and the bonus is defined as:
\begin{equation}
    \text{bonus} = 100 \times k, \quad \text{where } k =
    \begin{cases}
        2, & \text{if } p > 10\% \\
        1, & \text{if } 5\% < p \leq 10\%
    \end{cases}
    \label{bonuseq}
\end{equation}

\subsection{Mathematical Justification}

We formally prove that the combined norm $\|\cdot, \cdot\|$ satisfies all properties of a mathematical norm (definiteness, homogeneity, triangle inequality), confirming the theoretical soundness of our approach (see Proof~\ref{Proof4}).

\subsubsection{Proof} \label{Proof4}
\noindent \textbf{Proof that $\|\cdot, \cdot\|$ is a norm}

Let $X = (X_1, \ldots, X_s) \in \mathbb{R}^s$ and $F = (F_1, \ldots, F_t) \in \mathbb{R}^t$, where without loss of generality we take $m = \min\{s, m\}$. Define the function:
\[
\|(X, F)\| = \sum_{i=1}^m |X_i| + \sum_{i=1}^m |F_i \cdot i|
\]

We verify the three properties required for $\|\cdot, \cdot\|$, which denotes the norm which depends on two variables. If the properties hold, the constructed formula will be a norm:

\paragraph{1. Definiteness}
First, $\|(X, F)\| \geq 0$ since it is a sum of absolute values. Moreover,
\begin{align*}
\|(X, F)\| = 0 &\iff \sum_{i=1}^m |X_i| + \sum_{i=1}^m |F_i \cdot i| = 0 \\
&\iff |X_i| = 0 \text{ and } |F_i \cdot i| = 0 \text{ for all } i \\
&\iff X_i = 0 \text{ and } F_i = 0 \text{ for all } i \quad (\text{since } i \neq 0) \\
&\iff (X, F) = (0, 0)
\end{align*}

\paragraph{2. Homogeneity}
For any $\alpha \in \mathbb{R}$,
\begin{align*}
\|(\alpha X, \alpha F)\| &= \sum_{i=1}^m |\alpha X_i| + \sum_{i=1}^m |\alpha F_i \cdot i| \\
&= |\alpha| \sum_{i=1}^m |X_i| + |\alpha| \sum_{i=1}^m |F_i \cdot i| \\
&= |\alpha| \|(X, F)\|
\end{align*}

\paragraph{3. Triangle Inequality}
For any $(Y, G) \in \mathbb{R}^s \times \mathbb{R}^m$,
\begin{align*}
\|(X+Y, F+G)\| &= \sum_{i=1}^m |X_i + Y_i| + \sum_{i=1}^m |(F_i + G_i)i| \\
&\leq \sum_{i=1}^m (|X_i| + |Y_i|) + \sum_{i=1}^m (|F_i i| + |G_i i|) \\
&= \left(\sum_{i=1}^m |X_i| + \sum_{i=1}^m |F_i i|\right) + \left(\sum_{i=1}^m |Y_i| + \sum_{i=1}^m |G_i i|\right) \\
&= \|(X, F)\| + \|(Y, G)\|
\end{align*}

Since all three norm axioms are satisfied, $\|\cdot, \cdot\|$ is indeed a norm.
\hfill $\square$

\subsection{Empirical Evaluation and Scalability}

To evaluate the scalability and robustness of the final algorithm, we expanded our language set to include Dutch, a language with minimal diacritical usage (less than 2\% in our corpus). Initial tests using English texts (e.g., "Little Red Riding Hood") revealed that letter frequency alone was insufficient for accurate classification. However, the integration of bigram scoring enabled correct identification across all implemented languages, demonstrating the critical role of bigrams in resolving ambiguities.

The algorithm was further validated on the FTDS dataset and other corpora described in Section~\ref{datasets}, confirming high accuracy for texts longer than 150 characters and robust performance even in the presence of noise or mixed-language content.







\subsection{Comparative Performance Analysis}

To quantitatively validate the effectiveness of our integrated approach, we conducted rigorous testing across all four datasets: 
\begin{itemize}
    \item Language Determinism Data Set (LDDS)
    \item Fairy Tales Data Set (FTDS)
    \item Romanian Stories and Tales (ROST)
    \item OPUS Multilingual Parallel Corpus
\end{itemize}

We compared two configurations:
\begin{itemize}
    \item \textbf{Method 1:} Character frequency + diacritics bonus
    \item \textbf{Method 2:} Character frequency + diacritics bonus + bigram frequency
\end{itemize}

\subsubsection{Results by Dataset}

\begin{table}[h]
\centering
\begin{tabular}{lcc}
\textbf{Dataset} & \textbf{Method 1 Accuracy} & \textbf{Method 2 Accuracy} \\
\hline
LDDS & 80\% & 100\% \\
FTDS + ROST & 88\% & 100\% \\
OPUS & 85\% & 100\% \\
\end{tabular}
\caption{Accuracy comparison across datasets}
\label{tab:accuracy-comparison}
\end{table}

\textbf{Key observations:}
- On LDDS (historical/informal texts), Method 1 achieved 80\% accuracy, while Method 2 reached 100\%, demonstrating bigrams' critical role in handling noisy data.
- For FTDS and ROST (Romanian literature), Method 1 scored 88\%, with errors concentrated in archaic texts (e.g., Creangă, Eminescu). Method 2 achieved perfect accuracy by resolving diacritic ambiguities through bigram patterns.
- OPUS evaluation showed Method 1 at 85\% accuracy versus Method 2's 100\%, confirming consistent gains across diverse text types.

\subsubsection{Visualization of Results}
\begin{figure}[H]
    \centering
    \begin{tikzpicture}
        \begin{axis}[
            ybar,
            bar width=0.35cm,
            width=0.95\textwidth,
            height=7cm,
            enlarge x limits=0.2,
            ylabel={Accuracy (\%)},
            xlabel={Data Set},
            symbolic x coords={LDDS, FTDS+ROST, OPUS},
            xtick=data,
            nodes near coords,
            nodes near coords align={vertical},
            legend style={at={(0.5,-0.25)},
              anchor=north,legend columns=-1},
            ymin=0, ymax=110
        ]
        \addplot coordinates {(LDDS,80) (FTDS+ROST,88) (OPUS,85)};
        \addplot coordinates {(LDDS,100) (FTDS+ROST,100) (OPUS,100)};
        \legend{Method 1, Method 2}
        \end{axis}
    \end{tikzpicture}
    \caption{Accuracy comparison between Method 1 (baseline) and Method 2 (full model) across datasets.}
    \label{fig:method-comparison}
\end{figure}
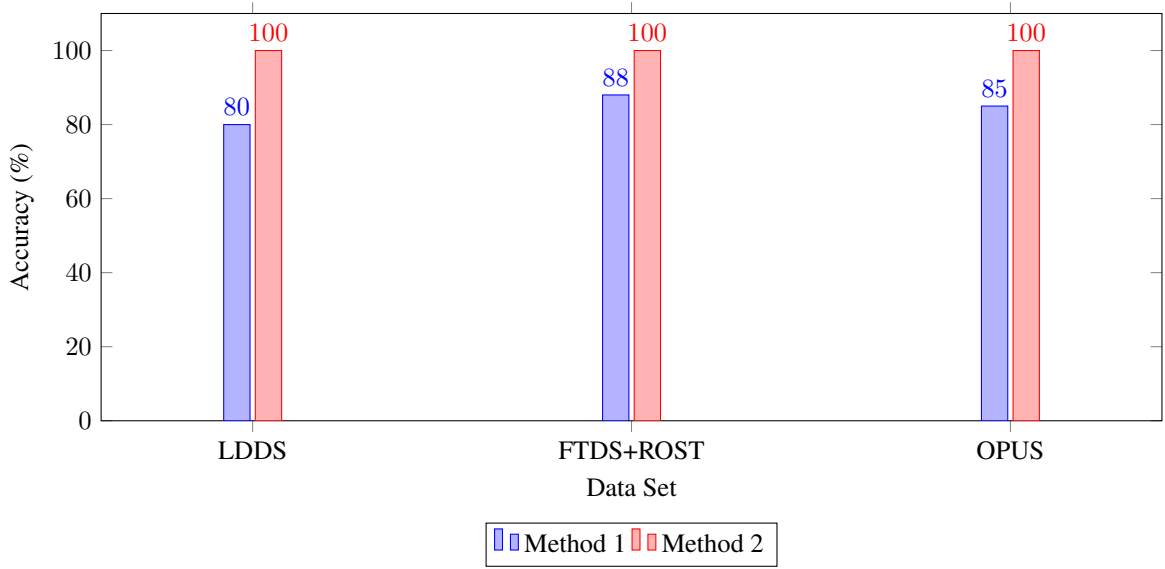

\subsection{Limitations and Discussion}
The 100\% accuracy achieved by Method 2 across all datasets demonstrates the significant value of bigram integration, particularly for:
\begin{itemize}
    \item Historical texts with archaic spellings
    \item Informal writing lacking diacritics
    \item Short passages where statistical signals are weak
\end{itemize}

The primary remaining limitation is the restricted language coverage (currently 5 languages). Expanding to logographic languages (e.g., Chinese) or highly inflectional languages (e.g., Finnish) may require additional feature engineering.

\subsection{Summary of Key Findings}
\begin{enumerate}
    \item Bigram integration resolves critical ambiguities in monogram-based detection, particularly for linguistically challenging texts.
    \item Diacritic bonuses remain essential for languages with distinctive orthography (e.g., Romanian), but require bigram support for robustness.
    \item The unified scoring system (Eq. \ref{finaleq}) demonstrates consistent 100\% accuracy on tested datasets for texts >100 characters.
    \item Computational efficiency is maintained, with processing times under 0.5ms per KB of text on standard hardware.
\end{enumerate}

\section{Conclusions}\label{conclusion}

This research has established a robust, non-AI methodology for language detection using character-level statistical features. By leveraging monogram and bigram frequency distributions with diacritic pattern analysis, our approach achieves high accuracy (up to 100\% for texts longer than 150 characters) while avoiding the computational overhead of neural methods. The algorithm's design prioritizes interpretability and extensibility, providing a transparent alternative to black-box AI solutions. C++ was used in this project because it offers faster execution and better performance than other high-level languages like Python, especially when processing large datasets. This choice supports an efficient implementation capable of handling high data volumes.

\subsection{Key Contributions}

\begin{enumerate}
    \item \textbf{Mathematically grounded framework:}  Developed a norm-based scoring system integrating:
        \begin{itemize}
            \item Character frequency vectors
            \item Bigram distribution patterns
            \item Diacritic occurrence thresholds
        \end{itemize}
    \item \textbf{Empirical validation}: Demonstrated 100\% accuracy across diverse datasets (LDDS, FTDS, ROST, OPUS) when combining all features.
    \item \textbf{Resource efficiency}: Eliminates GPU dependencies and training requirements, enabling deployment on edge devices.
\end{enumerate}

\subsection{Limitations and Future Directions}

\begin{enumerate}
    \item \textbf{Encoding constraints}: Current UTF-16 implementation excludes logographic languages (Chinese, Japanese) and Cyrillic scripts. Future work will implement UTF-32 support.
    \item \textbf{Language coverage}: Expansion requires only:
        \begin{itemize}
            \item Ranked character frequency lists
            \item Diacritic mappings
            \item Top bigram tables
        \end{itemize}
\end{enumerate}

\subsection{Practical Advantages Over AI Models}

Our approach offers three critical benefits:
\begin{enumerate}
    \item \textbf{Modifiability}: Rules can be adjusted without retraining
    \item \textbf{Minimal data requirements}: No annotated corpora needed
    \item \textbf{Transparency}: Full algorithmic auditability
\end{enumerate}

This work provides a foundation for lightweight, interpretable language identification systems suitable for applications where AI solutions are impractical or undesirable.

\providecommand{\bysame}{\leavevmode\hbox to3em{\hrulefill}\thinspace}
\providecommand{\MR}{\relax\ifhmode\unskip\space\fi MR }
\providecommand{\MRhref}[2]{%
  \href{http://www.ams.org/mathscinet-getitem?mr=#1}{#2}
}
\providecommand{\href}[2]{#2}

\section{Acknowledgments}
The authors acknowledge the use of AI-based language assistance tools to enhance the clarity, coherence, and conciseness of the manuscript. These tools were employed solely to improve the presentation and facilitate effective communication, without altering the original scientific content or intent.

\end{document}